\documentclass[runningheads]{llncs}
\usepackage[T1]{fontenc}

\usepackage[utf8]{inputenc} %
\usepackage[T1]{fontenc}    %
\usepackage{hyperref}       %
\usepackage{url}            %
\usepackage{booktabs}       %
\usepackage{amsfonts}       %
\usepackage{nicefrac}       %
\usepackage{microtype}      %
\usepackage{array}

\usepackage{times}
\usepackage{latexsym}
\usepackage{graphicx}
\usepackage{amssymb}
\usepackage{cleveref}

\usepackage{enumitem}
\usepackage{bm}
\usepackage{soul}
\usepackage{subfig}

\usepackage{footnote}

\hypersetup{
    colorlinks,
    linkcolor={red!50!black},
    citecolor={blue!50!black},
    urlcolor={blue!80!black}
}

\usepackage{graphicx}
\usepackage[numbers]{natbib}

\usepackage[normalem]{ulem} %
\usepackage{xcolor}

\let\svthefootnote\thefootnote
\newcommand\blankfootnote[1]{%
  \let\thefootnote\relax\footnotetext{#1}%
  \let\thefootnote\svthefootnote%
}

\begin{document}
\title{Multimodal Shannon Game with Images}
\author{
Vilém Zouhar$^{0\bigstar}$\orcidID{0000-0001-9874-2069}\and
Sunit Bhattacharya$^\bigstar$\orcidID{0000-0002-3271-4038} \and
Ondřej Bojar\orcidID{0000-0002-0606-0050}
}
\authorrunning{V. Zouhar et al.}
\institute{Institute of Formal and Applied Linguistics, Charles University
\\
\email{(zouhar,bhattacharya,bojar)@ufal.mff.cuni.cz}
\\
}
\maketitle              %
\blankfootnote{\hspace{-2mm}$^0$ Work done during stay at Charles University.}
\blankfootnote{\hspace{-2mm}$^\bigstar$ Equal contributions.}
\begin{abstract}

The Shannon game has long been used as a thought experiment in linguistics and NLP, asking participants to guess the next letter in a sentence based on its preceding context.
We extend the game by introducing an optional extra modality in the form of image information.
To investigate the impact of multimodal information in this game, we use human participants and a language model (LM, GPT-2).

We show that the addition of image information improves both self-reported confidence and accuracy for both humans and LM.
Certain word classes, such as nouns and determiners, benefit more from the additional modality information.
The priming effect in both humans and the LM becomes more apparent as the context size (extra modality information + sentence context) increases.
These findings highlight the potential of multimodal information in improving language understanding and modeling.

\end{abstract}

\noindent
\raisebox{-1mm}{\includegraphics[width=4mm]{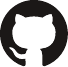}}
\hspace{0.1mm}
{
\fontsize{0.76em}{0.76em}\selectfont
Code: 
\href
{https://github.com/zouharvi/mmsg}
{\texttt{github.com/zouharvi/mmsg}}
\hfill
\raisebox{-1mm}{\includegraphics[width=4mm]{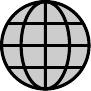}}
\hspace{0.1mm}
Annotation UI: \href
{https://vilda.net/s/mmsg/?uid=demo}
{\texttt{vilda.net/s/mmsg?uid=demo}}
}

\section{Introduction}

\begin{figure}[htpb]
\centering
\includegraphics[width=0.4\linewidth]{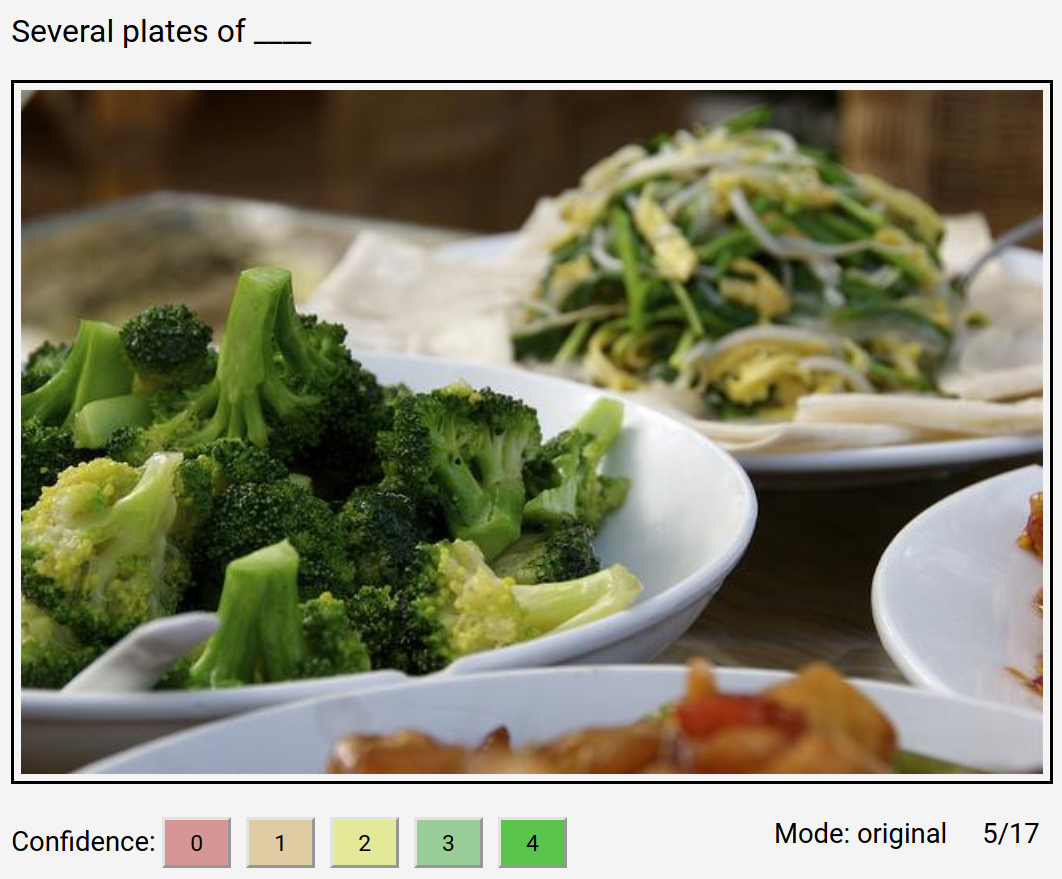}
\caption{Sentence ``\emph{Several plates of food are set on a table.}'' presented with an image. Given the first 3 words, the participant now has to think of the next word, rate their confidence and after ``food'' is revealed, self-evaluate how close they were.}
\label{fig:intro_original}
\end{figure}

The Shannon Game \cite{shannon1951prediction}\footnote{Not to be confused with Shannon's Switching Game.} is a well-known experiment from early 1950s that demonstrates the predictability of the English language.
Originally designed as a method to estimate the perplexity of a language, the game involves asking participants to predict the first letter of a text. Participants can choose from any of the 26 letters or space.
Upon making the guess, the correct character is revealed, and the participants are asked to guess the next (second) letter, and so on. 
When considering the game at the word level (\Cref{fig:intro_original}), it can be viewed as a variant of greedy autoregressive language modeling. As with autoregressive language modeling, the Shannon Game can be framed as the task of repeatedly predicting the probability of the next word given the previous context.

Numerous studies show that humans find it easier to process words that are probable given the context \cite{huang2011predictive,lupyan2015words,clark2013whatever}.
This phenomenon was widely studied in humans using the \textit{cloze procedure} \cite{taylor1953cloze}, where participants are presented with incomplete sentences and are asked to fill in the blanks using the context from both the left and right sides.
The Shannon Game can therefore also be seen as a version of the cloze task, where the word is to be predicted without the right-side context. Some studies have also investigated the impact of priming on word predictability using the cloze task \citep{kutas1984brain}.
However, these studies have been limited to a single, textual modality.

In this article, our aim is to address this gap and explore priming in multimodal conditions for both humans and language models (LM) like GPT-2 \citep{radford2019language}. We compare LM and human prediction capabilities in both text-only and multimodal settings.
To this end, we extend the Shannon Game to include an extra visual modality and investigate the relationship between self-reported confidence and accuracy of next-word prediction in humans and the LM. Furthermore, we relate the psycholinguistic concept of \emph{priming} to the neural language modeling concept of \emph{prompting}.

\begin{table}[htbp]
\centering
\begin{tabular}{r<{\hspace{2mm}}p{8.6cm}}
\toprule
\textbf{No image} & No extra information was shown and the participants could only use the left context. \\
\textbf{Original} & The full original image was shown.
\vspace{0.5mm}\\
\textbf{Labels  all} & The full original image was shown with bounding boxes and labels (\Cref{fig:labels_all}). \\
\textbf{Labels  crop} & The detected parts of the image were cropped and the snippets shown with labels (\Cref{fig:labels_crop}). \\
\textbf{Labels  text} & Only the list of labels of objects in the image was shown (\Cref{fig:labels_text}). \\
\bottomrule
\end{tabular}
\caption{Possible multimodality configurations.}
\label{tab:configs}
\end{table}

\section{Related Work}

Early research on the impact of contextual information on lexical prediction during reading relied on sentence prediction tasks \cite{fischler1979automatic,kleiman1980sentence}.
This concept was first introduced as a Shannon Game by Goodman \cite{goldman1958speech}, followed by the earliest versions involving images \cite{attneave1954some,barlow1961possible,kersten1987predictability}.
In fact, reading and sentence prediction have been compared to a \textit{psycholinguistic guessing game} \cite{goodman1969analysis,cairns1975lexical,goodman2014reading}.
We posit that a task like next word prediction in a sentence provides an interesting opportunity to study the impact of context in language processing and predictability.
With the exploration of predictive processing in reading \cite{wlotko2015time}, we can utilize these developments to design our experiment.

The effect of context is pervasive and present at multiple levels of processing \cite{willems2021context}. Previous fMRI studies \cite{mummery1999dual,rissman2003event} have demonstrated that the brain's response to a given word depends on the preceding linguistic context.
\cite{ames2015contextual} explore the impact of contextual information, in the form of visual data, on discourse comprehension and \cite{altmann2009incrementality} provide a comprehensive cognitive explanation of how visual context affects language processing, reporting that the ``eyes move toward whatever in the visual scene that unfolding word could refer to.''
Several psycholinguistically motivated studies \cite{barca2012unfolding, vanderwart1984priming} have investigated the role of general context in lexical prediction and how cross-modal priming (with images and text) works in lexical decision tasks.

However, these studies did not explicitly investigate semantic priming for a cloze task.
We attempt to do so in a cross-modal setting. Our aim is to explore the extent of semantic priming in a Shannon Game setting when priming is done using an image or information extracted from that image, in the direction of \cite{cho2021unifying}. Hence, the Multimodal Shannon Game with images can also be perceived as an autoregressive image captioning task, where the output is generated word-by-word, and its accuracy can be easily measured \cite{hossain2019comprehensive}.

In this direction, \cite{bhattacharya2022emmt} conducted an experiment on human participants with translation enhanced by image modality, which is parallel to our experiment with language modelling using the same modality. Finally, some researchers \cite{hladka2009designing,hladka2011attractive} have utilized the \emph{Games With a Purpose} methodology \cite{von2008designing} to frame tasks that are difficult for computers but relatively easy for humans as games. Similarly, we frame our experiment as a game that participants reportedly enjoy.

\section{Priming and Prompting}

Priming is a psychological and linguistic phenomenon where the presentation of a stimulus affects the processing of another stimulus in the future.
This effect has been widely studied in various contexts and has been defined as the facilitative effect of an encounter with a stimulus on subsequent processing of the same or a related stimulus \cite{tulving1982priming}.
One of the most important paradigms of priming is semantic priming, where the response to a stimulus is faster if it is preceded by something semantically related.
For example, the reaction to the word ``dog'' in a sentence would be faster if a semantically related prime, like ``cat'', were presented previously in the sentence \cite{meyer1971facilitation, shelton1992semantic}.

Prompting is a relatively new paradigm in neural language modeling where pretrained language models are trained to perform several downstream tasks by using an appropriate ``prompting function'' \cite{liu2021pre}.
In this paradigm, a pretrained language model is conditioned on extra information in the context, in addition to the previous words, to model $p(w_i| w_{<i}, C)$ i.e. the probability of predicting the next word given the previous words and the additional image context.

We use the Multimodal Shannon Game (MMSG) framework to assess whether semantic priming benefits autoregressive language models in the same way as it does humans.
Specifically, that the additional visual information (in whichever form) helps in the next word prediction task in the same way for both humans and LMs.
The results of this study contribute to our understanding of how multimodal information can be used to improve language modeling and documentation of the semantic priming effects.

\begin{figure*}
\centering

\begin{minipage}{\linewidth}
\subfloat[\emph{original}]{\includegraphics[width=0.4\linewidth]{img/annotation_original.png}\label{fig:original}}
\hfill
\subfloat[\emph{labels all}]{\includegraphics[width=0.4\linewidth]{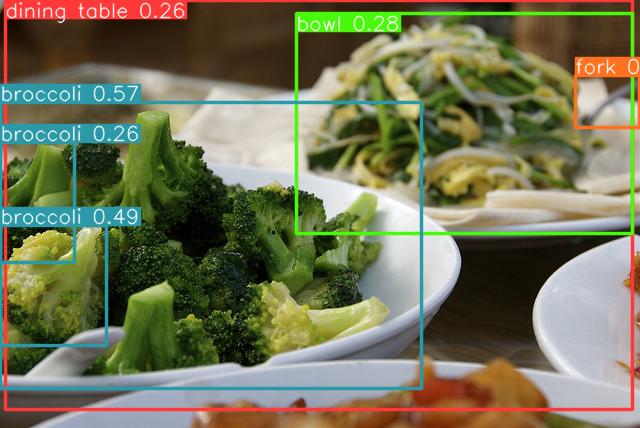}\label{fig:labels_all}}
\end{minipage}

\vspace{0.5cm}

\begin{minipage}{\linewidth}
\centering
\subfloat[\emph{labels crop}]{\includegraphics[width=0.4\linewidth]{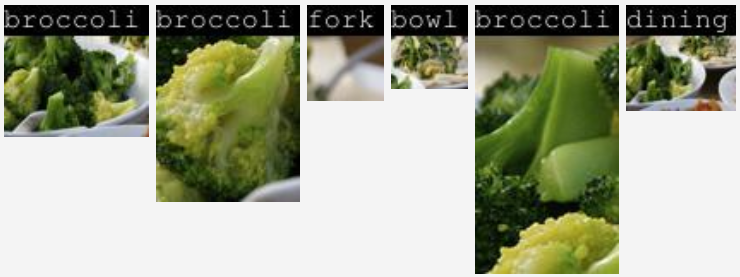}\label{fig:labels_crop}}
\hfill
\subfloat[\emph{labels text}]{\includegraphics[width=0.4\linewidth]{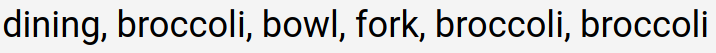}\label{fig:labels_text}}
\end{minipage}
\caption{The 4 configurations of multimodality for the same sentence (``\emph{Several plates of food are set on a table.}''). Given the first 3 words, the participant now has to think of the next word, rate their confidence and after \emph{food} is revealed, self-evaluate how accurate they were. The configuration \emph{no image} is not shown.}
\label{fig:modes}
\end{figure*}

\section{Experiment setup}

\paragraph{Methodology.}

The MMSG experiment consists of asking participants to predict the next word based on the previous (left) context, optionally given a related image information (see example in \Cref{fig:intro_original}).
We consider five configurations as described in \Cref{tab:configs}.
All participants saw each of the 17 sentences (listed in \Cref{sec:sentences}) with a randomly generated configuration.
\emph{No image} corresponds to vanilla autoregressive LM while \emph{Original} corresponds to a multimodal LM which also processes an image. 
\emph{Labels all}, \emph{labels crop} and \emph{labels text} correspond to pipelines that use an image object detector as an intermediate step.
Examples of configurations are shown in \Cref{fig:intro_original,fig:modes}.

\begin{table}[htbp]
{
\begin{enumerate}[noitemsep,left=0mm]
\item[0.] To be or not to be, that is the question
\item A girl with some food and drink at a table.
\item Man and son standing on the beach side with a self assembled kite.
\item A medium sized home kitchen with wood cabinets.
\item Several plates of food are set on a table.
\item A woman slashing down a snowy hill on skis.
\item An Adidas advertisement depicts a male and a female tennis player on the court.
\item A small brown teddy bear sitting on top of a box.
\item A macbook laptop next to a phone, backpack, and various books.
\item A person sits in a small boat on the water.
\item People watching an on screen presentation of a gentleman in a suit.
\item Food trucks are parked around small oval tables.
\item A couple of black cows standing on the top of a grassy hill.
\item A group of young children riding skis down a snow covered mountain.
\item Young children sharing a laptop in a messy room with several laptops, books, and papers.
\item A black sculpture of a torso is on the floor next to a TV.
\item United States President Barack Obama gives a speech in front of American and Russian flags.
\end{enumerate}
}
\caption{Sentences used in the experiment in fixed order.}
\label{sec:sentences}
\end{table}

\begin{table*}[htbp]
\centering
\begin{tabular}{lp{6.5cm}p{4.6cm}}
\toprule
\bf \# & \bf Confidence & \bf Accuracy \\
\midrule 
\textbf{0} & You have no idea about the next word.
& Could not be more wrong (wrong area and POS) \\
\textbf{1} & You know at least e.g. what part-of-speech the next word probably is.
& Very wrong but some aspects close (e.g. POS) \\
\textbf{2} & You know roughly what areas of words to expect.
& Wrong but the idea was roughly right \\
\textbf{3} & You know the next word or some variations of it.
& Very close (same area and POS) \\
\textbf{4} & You know the next word precisely.
& Exact match \\
\bottomrule
\end{tabular}
\caption{Description of the \textbf{confidence} and \textbf{accuracy} scale shown to the participants.}
\label{tab:conf_desc}
\end{table*}

\paragraph{Participants.}

We enrolled 24 volunteers from the academic environment, aged 24 to 40 years of various nationalities.
They were all non-native English speakers with advanced language proficiency (\href{https://en.wikipedia.org/wiki/Common_European_Framework_of_Reference_for_Languages}{C1 and C2 levels}).

\begin{figure}[htbp]
\begin{center}
\includegraphics[width=0.8\linewidth]{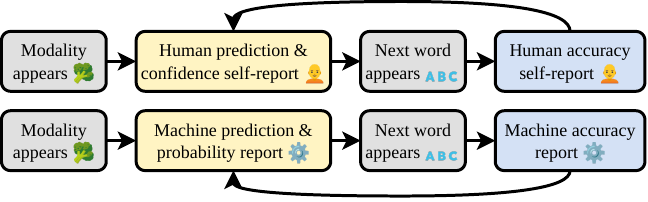}
\end{center}
\caption{Annotation pipeline for Multimodal Shannon Game with images. The loop ends when the end of sentence is reached.}
\label{fig:annotation_pipeline}
\end{figure}

\paragraph{Annotation environment.}

The annotation environment used in the MMSG experiment consists of a sequence of screens for each sentence.
Each screen starts with one of the five configurations, a blank ``\texttt{\_\_}'' cursor and the participant being asked to guess the first word and mark their confidence on a numeric scale (\Cref{tab:conf_desc}).
Upon pressing any of the five buttons, the actual next word is revealed and the participants are presented with a self-evaluation scale (\Cref{tab:conf_desc}).
Afterwards, they guess the next word and so on until the end of the sentence.
See the full instructions.\footnote{
Instructions:
In this experiment you're going to be predicting the next word in a sentence, starting with the first word.
Your task is to think about the next word (a specific word) and then click a number corresponding to how confident you are in your prediction.
Afterwards, the word is shown and you should evaluate how close your prediction was.
Some of the sentences may be accompanied by images, labelled images, a set of labels or snippets of items (you may need to scroll down to see all).
You should use these to improve your prediction.
The whole session should not last longer than 20-30 minutes.
Please take breaks only after you just finished your sentence, before clicking next sentence.
This is important as we are evaluating also the reaction times.
Do not close this window throughout the experiment as your progress would be lost.
}
The overall pipeline, for humans and LMs, is shown in \Cref{fig:annotation_pipeline}.
See \Cref{fig:demo_2b} in the appendix for the user interface thorough the whole pipeline.

\begin{figure*}
\includegraphics[width=0.33\linewidth]{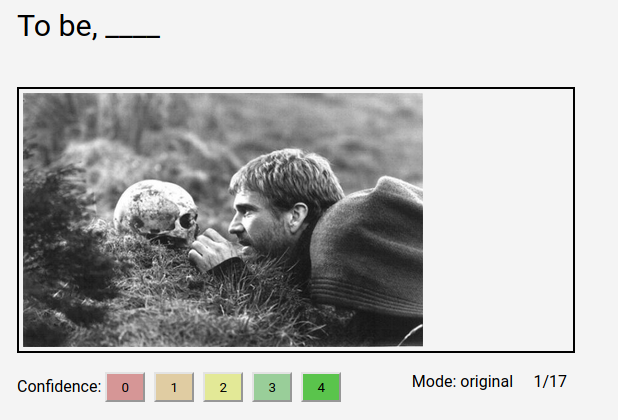}
\includegraphics[width=0.33\linewidth]{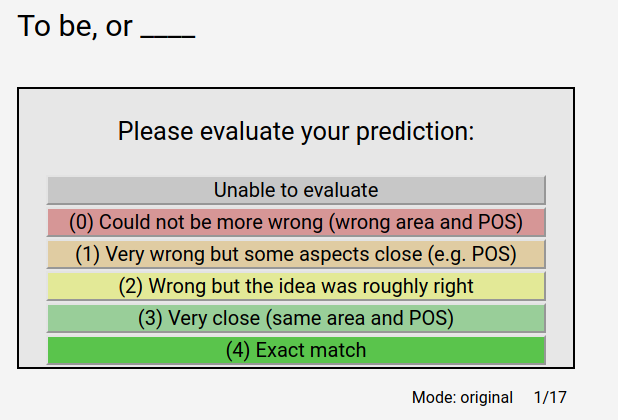}
\includegraphics[width=0.33\linewidth]{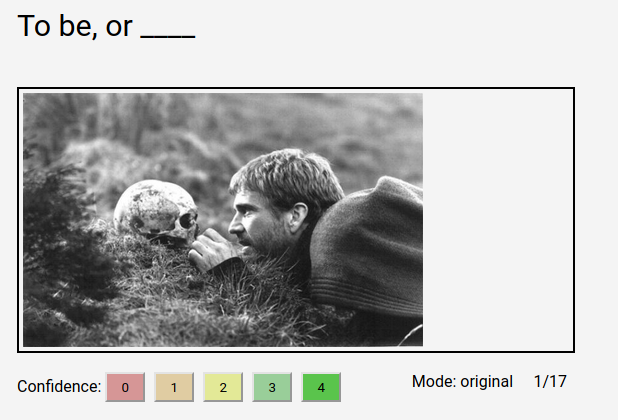}
\caption{prediction and self-evaluation of a single word based on the information of ``\textit{To be,}'' and the image.}
\label{fig:demo_2b}
\end{figure*}

The experiment was implemented as an web application, which allowed us to reach more participants at the cost of having no control over the environment.\footnote{The annotation environment was shown on various browser versions of the participants.}

\begin{figure*}[htbp]
\centering
\includegraphics[width=\linewidth]{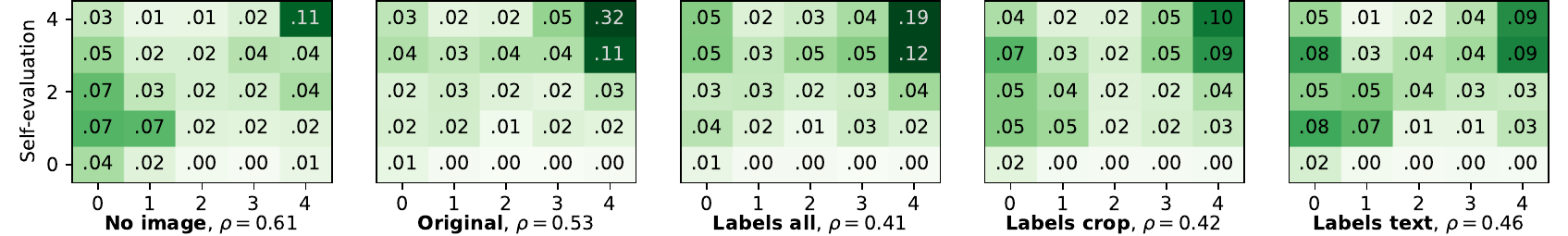}
\caption{Heatmap of confidence$\times$self-eval scores across configurations. The x-axis is the confidence score. Each cell reports the relative number of such judgements. Correlations ($\rho$) are Pearson's correlation coefficients between the confidence and self-eval scores.}
\label{fig:grade_conf}
\end{figure*}

\paragraph{Sentences.}

We selected 16 English sentences of length between 8 to 15 words.
This scale was chosen so that the participants are fully focused during the whole session (average of 25 minutes).
Furthermore, the smaller scale is required to have a more representative sample for each sentence + configuration tuple.
Note that this is not the natural distribution of the sentence length but desirable from an experiment design perspective to be able to compare phenomena across this variable.
Because they were taken from an image captioning dataset, some of the ``sentences'' are actually noun phrases without the main verb, which made the task more challenging for the participants.
The full list of the sentences is in \Cref{sec:sentences}.

We also added the sentence ``\emph{To be or not to be.}'' with an accompanying picture.
We assumed that the participants would easily recognize this sentence after the first few words and would continue with a sequence of high ratings.
This was meant to calibrate the participants' ratings and to introduce them to the task.

\begin{figure}[htbp]
\centering
\includegraphics[width=.7\linewidth]{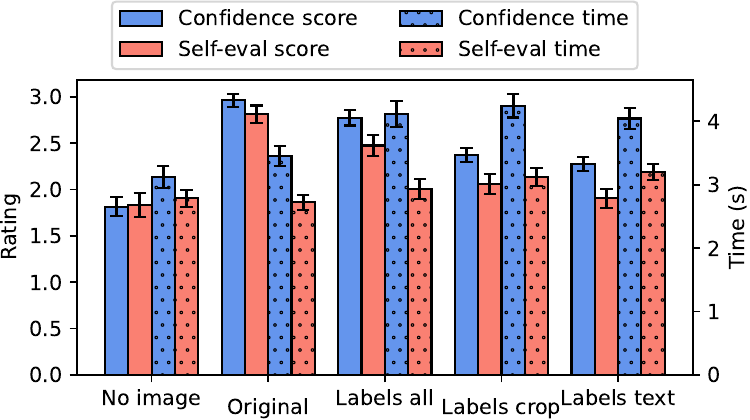}
\caption{Average confidence and self-eval scores and times. Confidence intervals are 95\% from t-distribution. Note the two separate y-axes for two kinds of quantities.}
\label{fig:basic_avgs}
\end{figure}

\section{Analysis}
\subsection{Effect of Configurations}

The confidence and accuracy averaged for each configuration are shown in \Cref{fig:basic_avgs}.
The \emph{original} configuration (where the entire image was shown to the participants) yielded both the highest confidence and self-evaluation scores while \emph{no image} configuration the lowest.
This shows that the participants were able to utilize the visual information.
When distilled to a set of labels (\emph{labels text}) or a series of pictures of individual objects extracted from the image (\emph{labels crop}), it still increased the confidence in their guesses with respect to the \emph{no image} configuration.

The difference in the self-reported accuracy and self-reported confidence for the configurations
\emph{labels text} and \emph{labels crop} is minimal. 
From a theoretical perspective, \emph{labels all} only added extra information in the form of bounding boxes and labels.
This had, unfortunately, a slightly detrimental effect in comparison to \emph{original}.
The participants agreed that the \emph{original} configuration was the easiest and that the \emph{labels all} was only distracting, in some cases obscuring an important part of the image and possibly suggested different synonyms than used in the sentence.

The distribution of confidence and self-eval scores is shown in \Cref{fig:grade_conf}, which also shows the bipolarity of the ratings.
Often the participants were either very sure and were correct (high scores) or the opposite (low scores) with few in-between.

\begin{figure}[htbp]
\centering
\includegraphics[width=.6\linewidth]{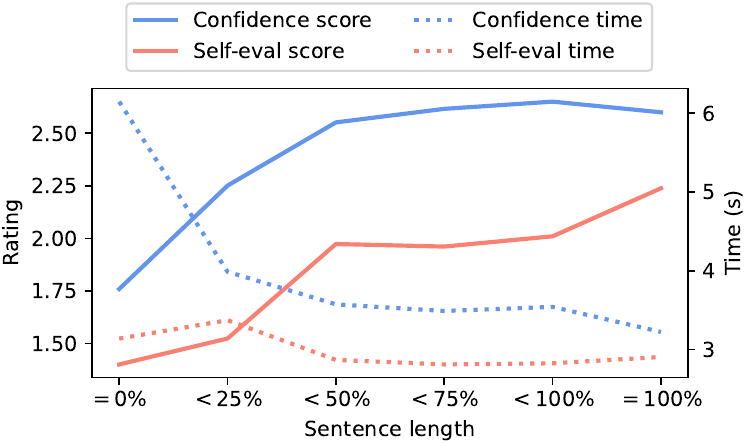}
\caption{Average prediction confidence and self-eval scores with reaction times with respect to the relative position in the sentence across all configurations.}
\label{fig:sent_len}
\end{figure}

\subsection{Effect of Word Position}

The first few words had naturally lower confidence and evaluation scores (accuracy), as shown in \Cref{fig:sent_len}.
This is expected on account of the space of all possible predicitions due to the limited available context.

For the first word, the participants used mostly one of two strategies: guessing an article or nothing at all.
The average confidence and self-evaluation for \emph{no image} was 1.19 and 0.48 and for \emph{original} was 2.14 and 2.16.
This is interesting as 10 out of 17 sentences begin with a determiner where the image should not help.

\begin{figure}[htbp]
\centering
\includegraphics[width=0.6\linewidth]{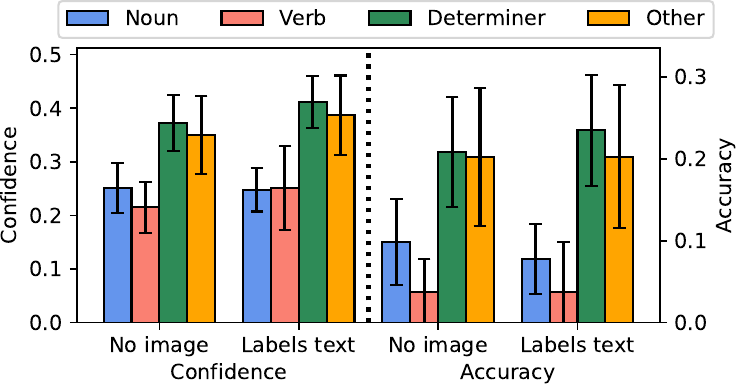}
\caption{Average POS prediction confidence and accuracy scores from GPT-2. Confidence intervals are 95\% from t-distribution.}
\label{fig:pos_avgs_gpt}
\end{figure}

\subsection{Effect of Part of Speech (POS)}

\begin{figure}[htbp]
\centering
\includegraphics[width=0.49\linewidth]{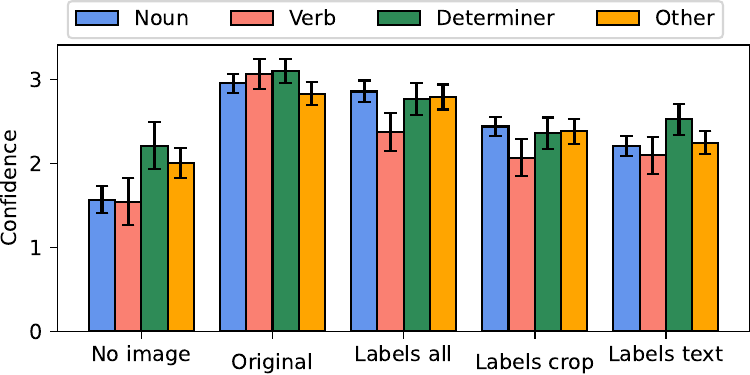} 
\includegraphics[width=0.49\linewidth]{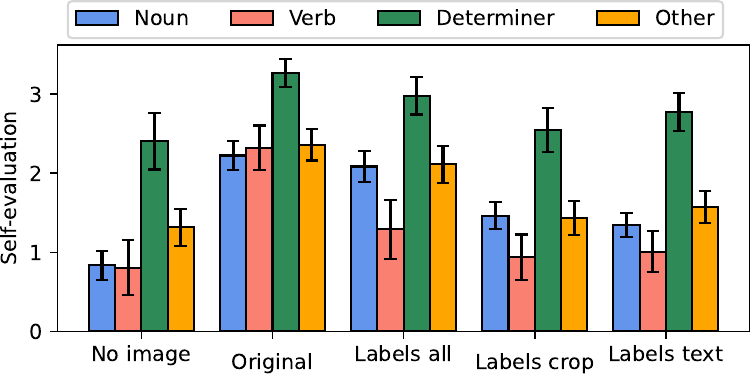}
\caption{Average POS prediction confidence and accuracy scores from humans. Confidence intervals are 95\% from t-distribution.}
\label{fig:pos_avgs}
\end{figure}

Naturally, some word classes are easier to predict than others.
This is shown in \Cref{fig:pos_avgs} where the users performed systematically better on determiners than other POS, like nouns.
Finally, for both accuracy and confidence, the \emph{no image} configuration yields the lowest values across all POS.
This is counterintuitive because the prediction of a determiner should be based on purely the syntactic properties of the left context and not the multimodality.
A possible explanation is the grammatical number disambiguation in the image.

Nouns are of interest because what the object labels represent in configurations like \emph{labels all}, \emph{labels crop} and \emph{labels text} are a sequence of nouns.
Nevertheless, For the nouns, we see the \emph{labels text} configuration yields the worst confidence and accuracy score from among other configurations with added modal information.
Even though we attempt at semantic priming of the nouns, the priming via text (\emph{labels text}) is comparatively less effective when analyzed with the confidence and accuracy scores of the human participants.

\subsection{LM Results}

We replicate the experiment on the GPT-2 language model \citep{radford2019language}.
In every step, for every word $w$ and model prediction $p$ (distribution across vocabulary), we use $\max p$ (maximum word probability) as the confidence of the model output.
The output, despite being a probability formally, is however not calibrated \citep{jiang2021can}.
Because GPT-2 is not a visual model, we consider only two configurations: \emph{no image} and \emph{labels text}.
\Cref{fig:pos_avgs_gpt} shows the results for GPT-2.
Slightly higher accuracy and confidence for the \emph{labels text} configuration show that the model is able to make use of the fusion to improve its prediction.
It exhibits some similar patterns to humans: lower confidence and accuracy for nouns and verbs and high for determiners.
The human-LM Pearson correlation coefficients for both confidence and accuracy decreases when we fuse in the labels (\Cref{tab:correlation}), suggesting different usage of the extra information in humans and LM.

\begin{table}[t]
\begin{center}
\begin{tabular}{lc>{\hspace{3mm}}c}
\toprule
& \emph{no image} & \emph{labels text} \\
\midrule
\textbf{Confidence} & 0.38 & 0.25 \\
\textbf{Accuracy} & 0.56 & 0.45 \\
\bottomrule
\end{tabular}
\end{center}
\caption{Pearson correlation coefficients (micro) between human annotators and GPT-2 predictions.}
\label{tab:correlation}
\end{table}

\section{Discussion}

Inspired by the 4 central questions (\emph{Why? What? How? When?}) about prediction in language processing proposed by \cite{huettig2015four}, we look at the results from the cognitive perspective.
We are primarily interested in the \emph{What?} questions i.e. what cues were relevant for the predictions and what language features are most affected with change in contextual cues.

Irrespective of the input modality, from the POS experiments it is evident that both the prediction confidence (anticipatory processing by \citep{kukona2014lexical}) and accuracy for verbs and nouns improve significantly with an informative multimodal context.
We also posit that the extra modality makes the models (and humans) more confident about the content of the sentence and that translates to the added confidence and accuracy of determiners.
For \cite{kukona2014lexical,kukona2011time} the modality was acoustic while our input modality was text and images.
Note that for GPT-2 the pattern of confidence and accuracy increasing with extra modal information does not fit perfectly with verbs.

In terms of the effects of priming in language models, \cite{sinclair2022structural,prasad2019using} use (syntactic and structural) priming to see how much language models are susceptible to priming effects.
Similarly, \cite{misra2020exploring} explored the effect of semantic priming in BERT.
Our formulation of the Multimodal Shannon Game establishes a way to effectively compare the priming effects in humans and LMs on the same benchmark which has not been attempted before.
We also find that the priming effect, as explored by us, gets more noticeable with additional context with autoregressive models, which contradits \cite{misra2020exploring}.
However, we do acknowledge that although they looked at the phenomenon of semantic priming, the methodology and the nature of stimuli used in \cite{misra2020exploring} is radically different. 

In summary, we see from the experiments that priming, the effects of which are well studied in humans can be related to prompting in large language models.

\section{Conclusion}
\label{Conclusion}

In this paper, we introduced the multimodal version of the Shannon Game and ran an experiment on human participants and we arrived at the following conclusions: We observed that the presence of any visual information positively influenced the confidence and accuracy of next-word prediction, with the full image configuration yielding the most significant improvements. We noted a mixed effect of both image configuration and word part-of-speech (POS) on prediction confidence and accuracy, indicating the complexity of multimodal interaction. 
Additionally, we found that the impact of priming became increasingly evident with a longer contextual span. 

Extending our study to the GPT-2 language model, we observed similar trends: GPT-2 benefitted from incorporating an additional modality, albeit with more variability. Word POS similarly influenced prediction confidence and accuracy in the GPT-2 model as in humans. Notably, the correlation of these metrics between human participants and GPT-2 decreased when an additional modality was introduced, suggesting differences in how humans and the model process inputs from the visual modality.

\section{Future work}
The space of extra modalities in the Shannon Game and cloze task is underexplored, consider e.g. video or audio.
The presented multimodal task could also be analyzed with standard psycholinguistic tools, such as EEG or eye-tracking.
Importantly, this experiment should be compared to multimodal language models and more recent models, which exhibit new, emergent, properties.

\section*{Limitations}

We focused on English, which may distribute information differently within sentences compared to other languages. A more morphologically rich language might be more predictable. All our participants were proficient, non-native English speakers. This however aligns with the fact that most English users are non-native.\footnote{\href{https://lemongrad.com/english-language-statistics/}{lemongrad.com/english-language-statistics}}
We used only GPT-2 models for our priming experiments and did not test larger models or different families, which might yield different patterns. Also, increasing our sample size could reduce the standard deviations in our results.

\section{Acknowledgements}
The work has been supported by the Ministry of Education, Youth and Sports
of the Czech Republic, Project No. LM2023062 (LINDAT/CLARIAH-CZ),
and by the grants 19-26934X (NEUREM3) of the Czech Science
Foundation, and 205-09/260698 (SVV) of Charles University.

\bibliographystyle{splncs04} %
\bibliography{misc/bibliography}

\begin{thebibliography}{10}
\providecommand{\url}[1]{\texttt{#1}}
\providecommand{\urlprefix}{URL }
\providecommand{\doi}[1]{https://doi.org/#1}

\bibitem{altmann2009incrementality}
Altmann, G.T., Mirkovi{\'c}, J.: Incrementality and prediction in human sentence processing. Cognitive science  \textbf{33}(4),  583--609 (2009), \url{https://onlinelibrary.wiley.com/doi/full/10.1111/j.1551-6709.2009.01022.x}

\bibitem{ames2015contextual}
Ames, D.L., Honey, C.J., Chow, M.A., Todorov, A., Hasson, U.: Contextual alignment of cognitive and neural dynamics. Journal of cognitive neuroscience  \textbf{27}(4),  655--664 (2015), \url{https://www.sciencedirect.com/science/article/pii/S2589004221003606}

\bibitem{attneave1954some}
Attneave, F.: Some informational aspects of visual perception. Psychological review  \textbf{61}(3), ~183 (1954), \url{https://psycnet.apa.org/record/1955-01960-001}

\bibitem{barca2012unfolding}
Barca, L., Pezzulo, G.: Unfolding visual lexical decision in time. PloS one  \textbf{7}(4),  e35932 (2012), \url{https://journals.plos.org/plosone/article?id=10.1371/journal.pone.0035932}

\bibitem{barlow1961possible}
Barlow, H.B., et~al.: Possible principles underlying the transformation of sensory messages. Sensory communication  \textbf{1}(01) (1961), \url{https://www.cnbc.cmu.edu/~tai/nc19journalclubs/Barlow-SensoryCommunication-1961.pdf}

\bibitem{bhattacharya2022emmt}
Bhattacharya, S., Kloudov{\'a}, V., Zouhar, V., Bojar, O.: {EMMT}: {A} simultaneous eye-tracking, 4-electrode {EEG} and audio corpus for multi-modal reading and translation scenarios. arXiv preprint arXiv:2204.02905  (2022), \url{https://arxiv.org/abs/2204.02905}

\bibitem{cairns1975lexical}
Cairns, H.S., Kamerman, J.: Lexical information processing during sentence comprehension. Journal of Verbal Learning and Verbal Behavior  \textbf{14}(2),  170--179 (1975), \url{https://www.sciencedirect.com/science/article/pii/S0022537175800636}

\bibitem{cho2021unifying}
Cho, J., Lei, J., Tan, H., Bansal, M.: Unifying vision-and-language tasks via text generation. In: International Conference on Machine Learning. pp. 1931--1942. PMLR (2021), \url{https://proceedings.mlr.press/v139/cho21a.html}

\bibitem{clark2013whatever}
Clark, A.: Whatever next? predictive brains, situated agents, and the future of cognitive science. Behavioral and brain sciences  \textbf{36}(3),  181--204 (2013), \url{http://apophenia.wdfiles.com/local--files/start/Clark_Whatever_NextDec2011.pdf}

\bibitem{fischler1979automatic}
Fischler, I., Bloom, P.A.: Automatic and attentional processes in the effects of sentence contexts on word recognition. Journal of verbal learning and verbal behavior  \textbf{18}(1),  1--20 (1979), \url{https://www.sciencedirect.com/science/article/pii/S0022537179905346}

\bibitem{goldman1958speech}
Goldman-Eisler, F.: Speech production and the predictability of words in context. Quarterly Journal of Experimental Psychology  \textbf{10}(2),  96--106 (1958), \url{https://journals.sagepub.com/doi/abs/10.1080/17470215808416261}

\bibitem{goodman1969analysis}
Goodman, K.S.: Analysis of oral reading miscues: {A}pplied psycholinguistics. Reading research quarterly pp. 9--30 (1969), \url{https://www.jstor.org/stable/747158}

\bibitem{goodman2014reading}
Goodman, K.S.: Reading: {A} psycholinguistic guessing game. In: Making sense of learners making sense of written language, pp. 115--124 (2014), \url{https://www.tandfonline.com/doi/pdf/10.1080/19388076709556976}

\bibitem{hladka2011attractive}
Hladk{\'a}, B., M{\'\i}rovsk{\`y}, J., Kohout, J.: An attractive game with the document:(im) possible? The Prague Bulletin of Mathematical Linguistics  \textbf{96}, ~5 (2011), \url{http://ufal.mff.cuni.cz/biblio/attachments/2011-vidova_hladka-p4229884754028822591.pdf}

\bibitem{hladka2009designing}
Hladk{\'a}, B., M{\'\i}rovsk{\`y}, J., Schlesinger, P.: Designing a language game for collecting coreference annotation. In: Proceedings of the Third Linguistic Annotation Workshop (LAW III). pp. 52--55 (2009), \url{https://aclanthology.org/W09-3008/}

\bibitem{hossain2019comprehensive}
Hossain, M.Z., Sohel, F., Shiratuddin, M.F., Laga, H.: A comprehensive survey of deep learning for image captioning. ACM Computing Surveys (CsUR)  \textbf{51}(6),  1--36 (2019), \url{https://dl.acm.org/doi/abs/10.1145/3295748}

\bibitem{huang2011predictive}
Huang, Y., Rao, R.P.: Predictive coding. Wiley Interdisciplinary Reviews: Cognitive Science  \textbf{2}(5),  580--593 (2011), \url{https://wires.onlinelibrary.wiley.com/doi/full/10.1002/wcs.142}

\bibitem{huettig2015four}
Huettig, F.: Four central questions about prediction in language processing. Brain research  \textbf{1626},  118--135 (2015), \url{https://www.sciencedirect.com/science/article/pii/S0006899315001146}

\bibitem{jiang2021can}
Jiang, Z., Araki, J., Ding, H., Neubig, G.: How can we know when language models know? on the calibration of language models for question answering. Transactions of the Association for Computational Linguistics  \textbf{9},  962--977 (2021), \url{https://direct.mit.edu/tacl/article/doi/10.1162/tacl_a_00407/107277/How-Can-We-Know-When-Language-Models-Know-On-the}

\bibitem{kersten1987predictability}
Kersten, D.: Predictability and redundancy of natural images. JOSA A  \textbf{4}(12),  2395--2400 (1987), \url{https://opg.optica.org/josaa/fulltext.cfm?uri=josaa-4-12-2395&id=2981}

\bibitem{kleiman1980sentence}
Kleiman, G.M.: Sentence frame contexts and lexical decisions: {S}entence-acceptability and word-relatedness effects. Memory \& Cognition  \textbf{8}(4),  336--344 (1980), \url{https://link.springer.com/article/10.3758/BF03198273}

\bibitem{kukona2014lexical}
Kukona, A., Cho, P.W., Magnuson, J.S., Tabor, W.: Lexical interference effects in sentence processing: {E}vidence from the visual world paradigm and self-organizing models. Journal of Experimental Psychology: Learning, Memory, and Cognition  \textbf{40}(2), ~326 (2014), \url{https://psycnet.apa.org/record/2013-40559-001}

\bibitem{kukona2011time}
Kukona, A., Fang, S.Y., Aicher, K.A., Chen, H., Magnuson, J.S.: The time course of anticipatory constraint integration. Cognition  \textbf{119}(1),  23--42 (2011), \url{https://www.sciencedirect.com/science/article/pii/S0010027710002933}

\bibitem{kutas1984brain}
Kutas, M., Hillyard, S.A.: Brain potentials during reading reflect word expectancy and semantic association. Nature  \textbf{307}(5947),  161--163 (1984), \url{https://www.nature.com/articles/307161a0}

\bibitem{liu2021pre}
Liu, P., Yuan, W., Fu, J., Jiang, Z., Hayashi, H., Neubig, G.: Pre-train, prompt, and predict: {A} systematic survey of prompting methods in natural language processing. arXiv preprint arXiv:2107.13586  (2021), \url{https://dl.acm.org/doi/full/10.1145/3560815}

\bibitem{lupyan2015words}
Lupyan, G., Clark, A.: Words and the world: {P}redictive coding and the language-perception-cognition interface. Current Directions in Psychological Science  \textbf{24}(4),  279--284 (2015), \url{https://journals.sagepub.com/doi/abs/10.1177/0963721415570732}

\bibitem{meyer1971facilitation}
Meyer, D.E., Schvaneveldt, R.W.: Facilitation in recognizing pairs of words: {Evidence} of a dependence between retrieval operations. Journal of experimental psychology  \textbf{90}(2), ~227 (1971), \url{https://psycnet.apa.org/record/1972-04123-001}

\bibitem{misra2020exploring}
Misra, K., Ettinger, A., Rayz, J.T.: Exploring {BERT}'s sensitivity to lexical cues using tests from semantic priming. arXiv preprint arXiv:2010.03010  (2020), \url{https://arxiv.org/abs/2010.03010}

\bibitem{mummery1999dual}
Mummery, C.J., Shallice, T., Price, C.: Dual-process model in semantic priming: {A} functional imaging perspective. Neuroimage  \textbf{9}(5),  516--525 (1999), \url{https://www.sciencedirect.com/science/article/pii/S1053811999904342}

\bibitem{prasad2019using}
Prasad, G., Van~Schijndel, M., Linzen, T.: Using priming to uncover the organization of syntactic representations in neural language models. arXiv preprint arXiv:1909.10579  (2019), \url{https://arxiv.org/abs/1909.10579}

\bibitem{radford2019language}
Radford, A., Wu, J., Child, R., Luan, D., Amodei, D., Sutskever, I.: Language models are unsupervised multitask learners  (2019), \url{https://cdn.openai.com/better-language-models/language_models_are_unsupervised_multitask_learners.pdf}

\bibitem{rissman2003event}
Rissman, J., Eliassen, J.C., Blumstein, S.E.: An event-related {fMRI} investigation of implicit semantic priming. Journal of cognitive neuroscience  \textbf{15}(8),  1160--1175 (2003), \url{https://direct.mit.edu/jocn/article/15/8/1160/3796/An-Event-Related-fMRI-Investigation-of-Implicit}

\bibitem{shannon1951prediction}
Shannon, C.E.: Prediction and entropy of printed english. Bell system technical journal  \textbf{30}(1),  50--64 (1951), \url{https://onlinelibrary.wiley.com/doi/abs/10.1002/j.1538-7305.1951.tb01366.x}

\bibitem{shelton1992semantic}
Shelton, J.R., Martin, R.C.: How semantic is automatic semantic priming? Journal of Experimental Psychology: Learning, memory, and cognition  \textbf{18}(6), ~1191 (1992), \url{https://psycnet.apa.org/record/1993-04339-001}

\bibitem{sinclair2022structural}
Sinclair, A., Jumelet, J., Zuidema, W., Fern{\'a}ndez, R.: Structural persistence in language models: {P}riming as a window into abstract language representations. Transactions of the Association for Computational Linguistics  \textbf{10},  1031--1050 (2022), \url{https://direct.mit.edu/tacl/article/doi/10.1162/tacl_a_00504/113019/Structural-Persistence-in-Language-Models-Priming}

\bibitem{taylor1953cloze}
Taylor, W.L.: “cloze procedure”: {A} new tool for measuring readability. Journalism quarterly  \textbf{30}(4),  415--433 (1953), \url{https://journals.sagepub.com/doi/full/10.1177/107769905303000400}

\bibitem{tulving1982priming}
Tulving, E., Schacter, D.L., Stark, H.A.: Priming effects in word-fragment completion are independent of recognition memory. Journal of experimental psychology: learning, memory, and cognition  \textbf{8}(4), ~336 (1982), \url{https://psycnet.apa.org/record/1982-31877-001}

\bibitem{vanderwart1984priming}
Vanderwart, M.: Priming by pictures in lexical decision. Journal of Verbal Learning and Verbal Behavior  \textbf{23}(1),  67--83 (1984), \url{https://www.sciencedirect.com/science/article/pii/S0022537184905097}

\bibitem{von2008designing}
Von~Ahn, L., Dabbish, L.: Designing games with a purpose. Communications of the ACM  \textbf{51}(8),  58--67 (2008), \url{https://dl.acm.org/doi/fullHtml/10.1145/1378704.1378719}

\bibitem{willems2021context}
Willems, R.M., Peelen, M.V.: How context changes the neural basis of perception and language. Iscience  \textbf{24}(5),  102392 (2021), \url{https://www.sciencedirect.com/science/article/pii/S2589004221003606}

\bibitem{wlotko2015time}
Wlotko, E.W., Federmeier, K.D.: Time for prediction? the effect of presentation rate on predictive sentence comprehension during word-by-word reading. Cortex  \textbf{68},  20--32 (2015), \url{https://www.sciencedirect.com/science/article/pii/S0010945215001057}

\end{thebibliography}

\end{document}